\documentclass{article}

\usepackage{microtype}
\usepackage{graphicx}
\usepackage{subcaption}
\usepackage{booktabs} 
\usepackage{hyperref}

\usepackage[accepted]{icml2026}

\usepackage{amsmath}
\usepackage{amssymb}
\usepackage{mathtools}
\usepackage{amsthm}

\usepackage[capitalize,noabbrev]{cleveref}

\theoremstyle{plain}

\theoremstyle{definition}

\theoremstyle{remark}

\usepackage{array}
\usepackage{tabularx}
\usepackage{url}
\usepackage{enumitem}
\usepackage{colortbl}
\usepackage{multirow}
\usepackage{bm}

\usepackage[textsize=tiny]{todonotes}

\icmltitlerunning{AgentPLM: Agentic Protein Language Models}

\begin{document}

\twocolumn[
  \icmltitle{AgentPLM: Agentic Protein Language Models with Reasoning-Augmented Decoding for Protein Sequence Design}




  \begin{icmlauthorlist}
    \icmlauthor{Sahil Rahman}{yyy}
    \icmlauthor{Maxx Richard Rahman}{comp}
  \end{icmlauthorlist}

  \icmlaffiliation{yyy}{Bedford College, London, United Kingdom}
  \icmlaffiliation{comp}{Saarland University, Saarbrücken, Germany}

  \icmlcorrespondingauthor{Sahil Rahman}{srahman@bedford.ac.uk}

  \icmlkeywords{Machine Learning, ICML}

  \vskip 0.3in
]



\printAffiliationsAndNotice{}  

\begin{abstract}
Protein language models (PLMs) are passive oracles: they generate sequences in a single forward pass with no mechanism to consult external biophysical feedback or redirect generation when a candidate violates thermodynamic or structural constraints. We introduce AgentPLM, which addresses this by equipping a pre-trained PLM with i) Reasoning-Augmented Decoding (RAD), which interleaves autoregressive generation with tool calls (ESMFold, FoldX, AutoDock Vina), and ii) Contrastive Agent Policy Optimisation (CAPO), a trajectory-level extension of direct preference optimisation that trains the policy end-to-end to learn when oracle feedback is informative rather than merely imitating high-fitness sequences. We evaluate AgentPLM on benchmark tasks spanning de novo enzyme design, antibody optimisation, thermostability, PPI interface design, and zero-shot fitness prediction with standardised oracle APIs and controlled sequence-identity splits. AgentPLM achieves state-of-the-art results with a gain in antibody top-10\% hit rate over the strongest passive baseline, providing mechanistic evidence of online error correction without explicit backtracking.
\end{abstract}

\section{Introduction}
\label{sec:introduction}
 
Protein language models (PLMs) have emerged as one of the most transformative tools in computational biology over the past five years.  Models such as ESM-2 \cite{lin2023esm2}, ProtTrans \cite{elnaggar2022prottrans}, and Ankh \cite{elnegeri2023ankh} demonstrate that transformer architectures pre-trained on large sequence databases can encode rich representations of evolutionary, structural, and functional information, powering state-of-the-art methods for structure prediction \cite{lin2023esm2}, inverse folding \cite{dauparas2022proteinmpnn}, and zero-shot fitness prediction \cite{notin2023proteingym}. The 2024 Nobel Prize in Chemistry, awarded in part for AlphaFold \cite{jumper2021alphafold}, crystallised the scientific community's recognition that machine learning has become indispensable to structural biology.  Yet a fundamental limitation pervades this generation of models, i.e., they are static oracles. A PLM generates sequences in a single forward pass, with no mechanism to interrogate the biophysical plausibility of intermediate steps, retrieve relevant experimental data, or adaptively redirect generation when a candidate sequence is predicted to be thermodynamically unstable or immunogenic. This passivity is in sharp contrast to how expert protein engineers actually work, iterating between computational predictions, wet-lab assays, and literature consultation in tightly coupled feedback loops \cite{arnold2018directed, yang2019machine}. The result is a fundamental mismatch, i.e., a PLM should implicitly encode all biophysical constraints in its parameters, yet its training signal (next-token prediction or masked token recovery on evolutionary sequences) does not directly supervise constraint satisfaction, making it systematically brittle for out-of-distribution design targets where the evolutionary record provides little guidance.
 
The analogous limitation in large language models for reasoning has been addressed. Chain-of-thought prompting \cite{wei2022chainofthought}, tool-augmented generation \cite{schick2023toolformer, qin2023toolllm}, and the ReAct framework \cite{yao2023react} have demonstrated that interleaving generation with structured reasoning steps and external tool calls dramatically improves performance on complex, multi-constraint tasks. ChemCrow \cite{bran2023chemcrow} applies this paradigm to chemistry and BioPlanner \cite{odonoghue2023bioplanner} to experimental protocol generation, while ProtAgent \cite{ghafarollahi2024protagents} uses a frozen GPT-4 backbone to orchestrate protein engineering tools through chain-of-thought planning. However, these systems treat the language model as a frozen planning module, preventing end-to-end training on domain-specific objectives and forcing the model to rely entirely on its parametric knowledge for sequence-level decisions. We argue that protein design is precisely the setting where this limitation is most damaging, i.e., generating a functional protein sequence requires satisfying simultaneous, non-additive constraints (thermostability, solubility, binding specificity, immunogenicity) that no single forward pass can jointly optimise, and the training signal needed to learn \emph{when} and \emph{how} to invoke structural oracles is available only through reinforcement on protein engineering objectives, not through general-purpose language modelling.
 
In this work we introduce AgentPLM, a framework that endows a pre-trained PLM with agentic capabilities by training it end-to-end to interleave autoregressive sequence generation with structured oracle invocations. We model each design step as a decision in a Partially Observable Markov Decision Process over the joint space of partial sequences and retrieved biophysical context, and we train the resulting policy with a novel Contrastive Agent Policy Optimisation (CAPO) objective that rewards trajectories leading to biophysically consistent, high-fitness sequences. The main contributions are:
 
\begin{itemize}[leftmargin=1.2em, itemsep=2pt, topsep=4pt]
 
  \item We propose Reasoning-Augmented Decoding (RAD), a decoding strategy that interleaves autoregressive sequence generation with structured tool calls, incorporating their outputs via a learned Tool Context Encoder (TCE) and Trajectory Memory Buffer (TMB) trained end-to-end on protein engineering objectives.
    
  \item We introduce Contrastive Agent Policy Optimisation (CAPO), which is a trajectory-level extension of direct preference optimisation that contrasts high-fitness trajectories with coherent oracle use against low-fitness or contradictory ones, teaching the model \emph{when} oracle feedback is informative rather than merely imitating high-fitness outputs.
 
  \item AgentPLM outperforms all the baselines on all the benchmark tasks, achieving a $2.79\times$ improvement in antibody top-10\% hit rate and $+34\%$ in normalised $k_{\mathrm{cat}}/K_{m}$ on enzyme design, confirming that gains arise from qualitatively different reasoning trajectories rather than additional compute.

\end{itemize}

\section{Related Works} 
\subsection{Protein Language Models and Generative Protein Design}
\label{sec:related:plm}
 
ESM-2 \cite{lin2023esm2}, ProtTrans \cite{elnaggar2022prottrans}, and Ankh \cite{elnegeri2023ankh} establish that masked and autoregressive transformers pre-trained on large sequence databases encode rich evolutionary and functional representations, with ESM-2 achieving state-of-the-art zero-shot fitness prediction on deep mutational scanning benchmarks \cite{notin2023proteingym}. Structure-conditioned design is addressed by ProteinMPNN \cite{dauparas2022proteinmpnn} via autoregressive inverse folding over backbone coordinates, while RFdiffusion \cite{watson2023rfdiffusion}, RFdiffusion-AA \cite{krishna2024rfaa}, and Chroma \cite{ingraham2023chroma} extend the paradigm to joint structure-sequence co-design through SE(3)-equivariant diffusion. For fitness-guided design, EvoProtGrad \cite{emami2023plug} climbs fitness landscapes via PLM-guided gradient descent over hundreds of mutation-selection cycles, and Tranception \cite{notin2022tranception} improves zero-shot scoring by retrieving homologous sequences at inference time. All of these models share a fundamental limitation: they generate sequences in a single forward pass or through oracle-free perturbations, so structural plausibility, thermodynamic stability, and binding specificity must be encoded entirely in parameters trained on evolutionary data, with no mechanism to observe constraint violations mid-generation and redirect accordingly \cite{huang2016coming,rahman2022, russ2020naturallike}.
 
\subsection{Agentic AI and Tool-Augmented Generation for Science}
\label{sec:related:agentic}
 
Toolformer \cite{schick2023toolformer} and ToolLLM \cite{qin2023toolllm} formalise the insertion of API call tokens into language model generation, while ReAct \cite{yao2023react} demonstrates that interleaving chain-of-thought reasoning \cite{wei2022chainofthought} with environment actions is mutually reinforcing. These frameworks have been applied to science via ChemCrow \cite{bran2023chemcrow} for organic synthesis and BioPlanner \cite{odonoghue2023bioplanner} for experimental protocol planning. In protein engineering, ProtAgent \cite{ghafarollahi2024protagents} is the closest prior work, using a frozen GPT-4 backbone to orchestrate ESMFold, FoldX, and literature search through chain-of-thought planning, but it shares two critical gaps with all prior tool-augmented systems: (i)~the language model cannot be trained end-to-end on protein objectives, preventing the tight coupling between oracle signals and residue-level generation that RAD achieves, and (ii)~no system integrates a literature oracle that fuses partial-sequence context with experimental knowledge at the token level, leaving residue-level mutagenesis knowledge inaccessible during generation \cite{kulmanov2022ProteinKG}. Prior RL approaches for protein design \cite{stanton2022bayesian, shanehsazzadeh2023denovo} apply reinforcement learning as a search strategy over discrete sequence space rather than as a policy-training objective over multi-step tool-use trajectories. CAPO extends direct preference optimisation \cite{rafailov2023dpo} to this trajectory-level contrastive setting, a formulation unexplored in either protein design or tool-augmented generation.

\begin{figure*}[t]
    \centering
    \includegraphics[width=0.95\textwidth]{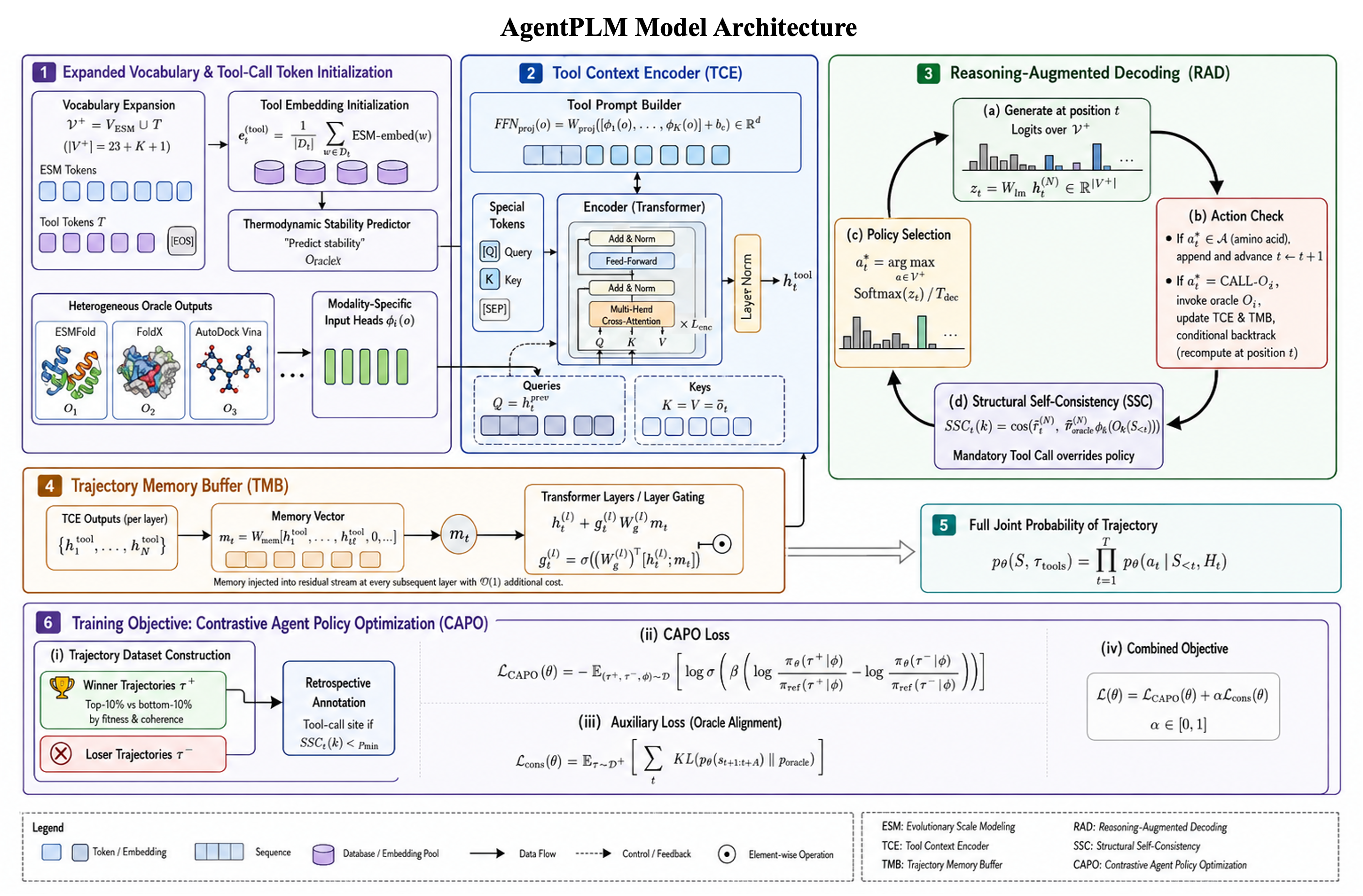}
    \caption{
    AgentPLM architecture comprising four modules, i) Expanded
  Vocabulary and Tool-Call Token Initialisation, ii) Trajectory Memory Buffer
  (TMB), iii) Reasoning-Augmented Decoding (RAD), and iv) trained end-to-end
  via the CAPO objective.
    }
    \label{fig:architecture}
\end{figure*}

\section{Problem Formulation}
\label{sec:problem}
 
\subsection{Protein Design as a Constrained Sequence Optimisation Problem}
\label{sec:problem:design}
Let $\mathcal{A}$ denote the amino-acid alphabet with $|\mathcal{A}|=20$, and let $\mathcal{A}^{*}=\bigcup_{L=1}^{\infty}\mathcal{A}^{L}$ be the set of all finite sequences. A protein design task is a tuple $\mathcal{T}=\langle f,\,\mathcal{C},\,S_{\mathrm{ref}}\rangle$, where $f:\mathcal{A}^{L}\!\to\!\mathbb{R}$ is a \emph{fitness function} (e.g.\ catalytic efficiency $k_{\mathrm{cat}}/K_{m}$, binding affinity $K_{D}$, or melting temperature $T_{m}$); $\mathcal{C}=\{c_{j}:\mathcal{A}^{L}\!\to\!\mathbb{R}\}_{j=1}^{J}$ is a set of \emph{biophysical constraint functions} (e.g.\ prescribed secondary-structure fractions, active-site geometry tolerances, immunogenicity thresholds), and $S_{\mathrm{ref}}\in\mathcal{A}^{L}$ is an optional reference (wild-type) sequence providing evolutionary context. The objective is:

\begin{equation}
\begin{aligned}
    S^{*} &= \arg\max_{S\in\mathcal{A}^{L}}\; f(S) \\
    &\quad\text{subject to}\quad c_{j}(S)\leq \varepsilon_{j},\quad \forall j\in[J].
\end{aligned}
\label{eqn:design_objective}
\end{equation}
  
where $\varepsilon_{j}$ are task-specific tolerances. In practice neither $f$ nor $c_{j}$ are analytically tractable, but can be estimated through a set of computational \emph{oracles} $\mathcal{O} \;=\; \{O_{k}:\mathcal{A}^{*}\!\to\!\mathbb{R}^{d_{k}}\}_{k=1}^{K}$, where $d_{k}$ is the output dimensionality of oracle $k$ (e.g.\ $O_{\mathrm{ESMFold}}$ returns per-residue pLDDT scores and Cartesian coordinates; $O_{\mathrm{FoldX}}$ returns a scalar $\Delta\Delta G$).

\subsection{Limitations of Passive Protein Language Models}
\label{sec:problem:passive}
 
A standard autoregressive PLM parameterises the sequence distribution as $p_{\theta}(S) \;=\; \prod_{i=1}^{L} p_{\theta}(s_{i}\mid s_{<i})$, where the parameters $\theta$ are learned by minimising next-token prediction loss on large sequence databases.  During inference the model generates the entire sequence in a single unidirectional pass, we call this a \emph{passive trajectory}:
\begin{equation} 
  \tau_{\mathrm{passive}} \;=\;
  (S_{\emptyset} \;\to\; S_{1} \;\to\; \cdots \;\to\; S_{L}),
  \label{eq:passive_traj}
\end{equation} 
where $S_{t}=(s_{1},\ldots,s_{t})$ is the partial sequence after $t$ amino acids have been committed.
 
\paragraph{Structural blindness.}
At each step $t$, the model selects $s_{t}\sim p_{\theta}(s_{t}\mid s_{<t})$ without observing $\{O_{k}(S_{t})\}_{k=1}^{K}$. This is a fundamental mismatch, i.e., all biophysical constraints should be implicitly encoded in $\theta$, yet the training signal (cross-entropy on evolutionary sequences) does not directly supervise constraint satisfaction. Formally, define the \emph{constraint violation} of a passive trajectory as $ \mathcal{V}(\tau_{\mathrm{passive}}) \;=\; \sum_{j=1}^{J}\max\!\bigl(0,\;c_{j}(S_{L})-\varepsilon_{j}\bigr).$ Because $c_{j}$ is never observed during generation, $\mathbb{E}[\mathcal{V}(\tau_{\mathrm{passive}})]$ cannot be minimised online. It can only be reduced by increasing model capacity, a strategy that is both computationally expensive and brittle for out-of-distribution targets.
 
\paragraph{Epistatic blindness.}
Long-range epistasis (where the fitness contribution of residue $s_{i}$ depends non-additively on a distant residue $s_{j}$) is systematically underestimated by autoregressive factorisation.  An oracle call at position $i$ that returns $O_{\mathrm{FoldX}}(S_{\leq i})$ can expose such dependencies before positions $j>i$ are committed, enabling corrective decisions.

\subsection{Agentic Protein Design}
\label{sec:problem:pomdp}
 
We reformulate protein design as a Partially Observable Markov Decision Process (POMDP) $\mathcal{M}=\langle\mathcal{X},\mathcal{A}^{+},\mathcal{O}, T,R,\gamma\rangle$:
 
\medskip
\noindent\textbf{State space.}\quad
$\mathcal{X} = \mathcal{A}^{*}\times\mathcal{H}$, where $\mathcal{H}=\bigl(\mathcal{A}^{*}\times[K]\times\mathbb{R}^{*}\bigr)^{*}$ is the space of tool-call histories.  The state at step $t$ is $x_{t}=(S_{\leq t}, H_{t})$, with $H_{t}=\{(S_{\leq t_{m}}, k_{m}, o_{m})\}_{m=1}^{M_{t}}$ recording the partial sequence, oracle index, and oracle output of each past tool call.
 
\medskip
\noindent\textbf{Augmented action space.}\quad
$\mathcal{A}^{+} \;=\; \mathcal{A} \;\cup\; \mathcal{T}, \qquad \mathcal{T} \;=\; \{\textsc{call-}O_{k}\}_{k=1}^{K}\;\cup\;\{\textsc{eos}\}$, where $\mathcal{A}$ is the amino-acid vocabulary, $\mathcal{T}$ is the tool-call vocabulary (one token per oracle plus an end-of-sequence token).
 
\medskip
\noindent\textbf{Observation function.}\quad
When $a_{t}\in\mathcal{T}\setminus\{\textsc{eos}\}$, the environment returns the oracle output $o_{t}=O_{k}(S_{\leq t})$, which is appended to $H_{t}$. When $a_{t}\in\mathcal{A}$, the observation is a null token and $H_{t}$ is unchanged.
 
\medskip
\noindent\textbf{Transition.}\quad
\begin{equation}
  x_{t+1} =
  \begin{cases}
    (S_{\leq t}\!\cdot\! a_{t},\; H_{t}) & \text{if } a_{t}\in\mathcal{A},\\
    (S_{\leq t},\; H_{t}\,\cup\,\{(S_{\leq t},k,o_{t})\}) & \text{if }
      a_{t}=\textsc{call-}O_{k}.
  \end{cases}
  \label{eq:transition}
\end{equation}
The tool calls do not advance the sequence index $t$. The agent re-scores position $t$ conditioned on the enriched history.
 
\medskip
\noindent\textbf{Reward.}\quad
$R(\tau) \;=\; f(S_{L}) \;-\; \lambda\,\bigl|\{t : a_{t}\in\mathcal{T}\}\bigr| \;-\; \mu\,\mathcal{V}(\tau)$, where $\lambda>0$ penalises excessive oracle calls (latency cost) and $\mu>0$ penalises biophysical constraint violations.  The agent's policy $\pi_{\theta}(a_{t}\mid x_{t})$ jointly models amino-acid selection and tool-invocation decisions.
 
\medskip
\noindent
A fundamental distinction from the passive PLM formulation in (Eq.~\ref{eq:passive_traj}) is that the agent can \emph{interrupt} generation to query oracles, incorporate their outputs into its context, and \emph{redirect} subsequent residue choices accordingly.  This transforms the one-shot constrained optimisation in Eq.~\eqref{eqn:design_objective} into a sequential decision problem where biophysical feedback is available online.


\section{AgentPLM Architecture}
\label{sec:architecture}
 
As shown in Figure~\ref{fig:architecture}, we propose AgentPLM consists of four modules, each addressing a distinct aspect of agentic sequence design.  
 
\subsection{Expanded Vocabulary and Tool-Call Token Initialisation}
\label{sec:arch:vocab}
 
\paragraph{Expanded vocabulary.}
The standard ESM-2 vocabulary $\mathcal{V}_{\mathrm{ESM}}=\mathcal{A}\cup \{\textsc{mask},\textsc{cls},\textsc{eos}\}$ (size $|\mathcal{A}|+3=23$) is extended to $\mathcal{V}^{+} \;=\; \mathcal{V}_{\mathrm{ESM}} \;\cup\; \mathcal{T}, \qquad |\mathcal{V}^{+}| = 23 + |\mathcal{T}|$, where $|\mathcal{T}|=K+1$ (one token per oracle plus \textsc{eos}).  The embedding matrix $\mathbf{E}\in\mathbb{R}^{|\mathcal{V}^{+}|\times d_{m}}$ is initialised as
\begin{equation}
  \mathbf{e}_{k}^{(\mathrm{tool})}
  \;=\; \frac{1}{|\mathcal{D}_{k}|}\sum_{w\in\mathcal{D}_{k}}
        \mathrm{ESM\text{-}embed}(w),
  \quad k\in[K],
  \label{eq:tool_embed_init}
\end{equation}
where $\mathcal{D}_{k}$ is a short natural-language description of the oracle $O_{k}$ (e.g. predict the change of thermodynamic stability after mutation for $O_{\mathrm{FoldX}}$) and $\mathrm{ESM\text{-}embed}(w)$ denotes the embedding of the ESM-2 sub-word in token $w$.  This semantically grounded initialisation avoids random initialisation of tool tokens and ensures they reside in the same representational manifold as amino-acid tokens from the outset of training. Existing tool-augmented LLMs like Toolformer~\cite{schick2023toolformer} and ToolLLM~\cite{qin2023toolllm} initialise special tokens from scratch, requiring large-scale fine-tuning to place them in the model's representational space. Our biologically motivated initialisation from Eq.~\eqref{eq:tool_embed_init} dramatically reduces the number of gradient steps required for the model to learn \emph{when} to invoke each oracle.

\paragraph{Tool Context Encoder (TCE).}
\label{sec:arch:tce}
 
Oracle outputs are heterogeneous. ESMFold returns a tensor of per-residue pLDDT scores and $C_{\alpha}$ coordinates $\mathbf{O}_{\mathrm{fold}}\in\mathbb{R}^{L\times 4}$, FoldX returns a scalar $\Delta\Delta G\in\mathbb{R}$, AutoDock Vina returns a binding score vector $\mathbf{v}\in\mathbb{R}^{9}$. A unified encoder is needed to project all these modalities into the model's latent space $\mathbb{R}^{d_{m}}$ ($d_{m}=1280$ for ESM-2 650M).
 
Given oracle output $o_{t}\in\mathbb{R}^{d_{k}}$ and the last hidden state $\mathbf{h}_{t}^{\mathrm{prev}}\in\mathbb{R}^{d_{m}}$ immediately preceding the tool call, the TCE computes a \emph{context embedding} via cross-attention: $\bar{\mathbf{o}}_{t} = \mathrm{FFN}_{\mathrm{proj}}(o_{t}) \in \mathbb{R}^{d_{m}}, \mathbf{h}_{t}^{\mathrm{tool}} = \mathrm{TCE}\!\left(o_{t};\,\mathbf{h}_{t}^{\mathrm{prev}}\right) = \mathrm{LayerNorm}\!\left( \mathbf{h}_{t}^{\mathrm{prev}} + \mathrm{Attn}\!\left( Q=\mathbf{h}_{t}^{\mathrm{prev}},\; K = V =\bar{\mathbf{o}}_{t} \right) \right),$ where $\mathrm{FFN}_{\mathrm{proj}}$ is a two-layer feed-forward network that maps oracle outputs of arbitrary dimension $d_{k}$ to $d_{m}$, using modality-specific input heads: $  \mathrm{FFN}_{\mathrm{proj}}(o) = \mathbf{W}_{2}\,\sigma\!\left(\mathbf{W}_{1}\, \phi_{k}(o) + \mathbf{b}_{1}\right)+\mathbf{b}_{2}, \quad \mathbf{W}_{1}\!\in\!\mathbb{R}^{2d_{m}\times d_{k}},\; \mathbf{W}_{2}\!\in\!\mathbb{R}^{d_{m}\times 2d_{m}},$with $\phi_{k}$ being a modality-specific pre-processing function:

\begin{equation}
  \phi_{k}(o) =
  \begin{cases}
    \mathrm{vec}(\mathbf{O}_{\mathrm{fold}}) & k=\mathrm{ESMFold},\\
    [\Delta\Delta G,\; \mathbf{1}^{\top}\mathbf{1}] & k=\mathrm{FoldX},\\
    \mathbf{v} & k=\mathrm{Vina}.
  \end{cases}
  \label{eq:phi}
\end{equation}

\subsection{Trajectory Memory Buffer (TMB)}
\label{sec:arch:tmb}
 
A single design trajectory may invoke up to $B_{\max}=8$ tool calls. Naively including all tool-call histories in the Transformer's context window would incur $\mathcal{O}(L^{2})$ attention cost proportional to the number of retrieved tokens.  The TMB provides a constant-size compressed episodic memory that conditions generation on the full tool-call history at $\mathcal{O}(1)$ additional cost.
 
Let $\{\mathbf{h}_{t_{1}}^{\mathrm{tool}},\ldots,\mathbf{h}_{t_{M}}^{\mathrm{tool}}\}$ be the sequence of TCE outputs for the $M$ tool calls in the current trajectory (with $M\leq B_{\max}$).  The TMB produces a memory vector $   \mathbf{m}_{t} \;=\; \mathbf{W}_{\mathrm{mem}}\, \Bigl[ \mathbf{h}_{t_{1}}^{\mathrm{tool}};\; \mathbf{h}_{t_{2}}^{\mathrm{tool}};\; \cdots;\; \mathbf{h}_{t_{M}}^{\mathrm{tool}};\; \mathbf{0}_{(B_{\max}-M)\,d_{m}} \Bigr], \mathbf{W}_{\mathrm{mem}} \\ \in\mathbb{R}^{d_{m}\times B_{\max}d_{m}}, $ where $[\,;\,]$ denotes concatenation and zero-padding ensures a fixed input dimension of $B_{\max}d_{m}$ regardless of the number of tool calls made so far. The memory vector $\mathbf{m}_{t}$ is injected into the model's residual stream at every subsequent Transformer layer via a learned \emph{memory gating} mechanism: $\tilde{\mathbf{h}}_{t}^{(\ell)} \;=\; \mathbf{h}_{t}^{(\ell)} \;+\; g_{t}^{(\ell)}\,\mathbf{W}_{\mathrm{gate}}^{(\ell)}\,\mathbf{m}_{t}, \quad g_{t}^{(\ell)} = \sigma\!\left(\mathbf{w}_{g}^{(\ell)\top} [\mathbf{h}_{t}^{(\ell)};\mathbf{m}_{t}]\right),$ where $g_{t}^{(\ell)}\in[0,1]$ is a scalar gate computed from the current hidden state and memory, $\mathbf{W}_{\mathrm{gate}}^{(\ell)}\in\mathbb{R}^{d_{m}\times d_{m}}$ is a layer-specific projection, and $\sigma$ is the sigmoid function. The gating ensures the memory contribution is \emph{learned} and layer-dependent, rather than being added uniformly.

\subsection{Reasoning-Augmented Decoding (RAD)}
\label{sec:arch:rad}
RAD is the inference-time procedure that orchestrates the POMDP policy
$\pi_{\theta}(a_{t}\mid x_{t})$ over the augmented action space
$\mathcal{A}^{+}$.
 
\paragraph{Decoding loop.}
At each position $t$, the model computes logits over $\mathcal{V}^{+}$: $  \mathbf{z}_{t} \;=\; \mathbf{W}_{\mathrm{lm}}\,\tilde{\mathbf{h}}_{t}^{(N)} \;\in\;\mathbb{R}^{|\mathcal{V}^{+}|}, $ where $\tilde{\mathbf{h}}_{t}^{(N)}$ is the memory-gated output of the final Transformer layer.  The policy then selects:
\begin{equation}
  a_{t}^{*} \;=\; \arg\max_{a\in\mathcal{V}^{+}}\;
  \frac{\exp(z_{t,a}/T_{\mathrm{dec}})}
       {\sum_{a'}\exp(z_{t,a'}/T_{\mathrm{dec}})},
  \label{eq:rad_action}
\end{equation}
where $T_{\mathrm{dec}}$ is the decoding temperature.  If $a_{t}^{*}\in\mathcal{A}$, the amino acid is appended and $t\leftarrow t+1$.  If $a_{t}^{*}=\textsc{call-}O_{k}$, the oracle is invoked, the TCE updates the representation, and the TMB is updated without advancing $t$.  The logits are then recomputed at position $t$ with the enriched representation, effectively implementing a conditional backtrack step.
 
\paragraph{Structural self-consistency score.}
Before finalising each tool-call decision, we compute a \emph{structural self-consistency} (SSC) score that measures whether the oracle feedback is consistent with the model's internal belief about the partial sequence: $\mathrm{SSC}_{t}(k) \;=\; \cos\!\Bigl( \tilde{\mathbf{h}}_{t}^{(N)},\; \mathbf{W}_{\mathrm{oracle}}^{(k)}\, \phi_{k}(O_{k}(S_{\leq t})) \Bigr) \;\in\;[-1, 1]$, where $\mathbf{W}_{\mathrm{oracle}}^{(k)}\in\mathbb{R}^{d_{m}\times d_{k}}$ is a learned projection that maps oracle $k$'s output into the model's hidden space. A low SSC score (below threshold $\rho_{\min}=-0.2$) triggers a mandatory tool call at position $t$, overriding the policy's action if necessary:
\begin{equation}
  a_{t} \;=\;
  \begin{cases}
    \textsc{call-}O_{k^{*}} & \min_{k}\mathrm{SSC}_{t}(k) < \rho_{\min} 
                              \text{ and } \delta_{t}\geq\Delta_{\min},\\
    a_{t}^{*} & \text{otherwise},
  \end{cases}
  \label{eq:rad_override}
\end{equation}
where $k^{*}=\arg\min_{k}\mathrm{SSC}_{t}(k)$ is the oracle with the lowest
self-consistency (most surprising prediction) and $\delta_{t}$ is the number
of amino acids generated since the last tool call.
 
\paragraph{Budget constraints.}
To prevent degenerate tool-call loops, RAD enforces two hard constraints: (1)~a minimum inter-call gap $\delta_{t}\geq\Delta_{\min}=10$ amino acids, and (2)~a maximum tool-call budget $B_{\max}=8$ per sequence.  Together these ensure the generation process terminates in $\mathcal{O}(L+B_{\max}\cdot\bar{T})$ time, where $\bar{T}$ is the mean oracle latency.
 
\paragraph{Joint probability of a trajectory.}
The full conditional distribution over sequences and tool-call trajectories is:
\begin{equation}
  p_{\theta}(S, \tau_{\mathrm{tools}})
  \;=\;
  \prod_{t=1}^{L}
  p_{\theta}\!\left(a_{t}\;\Big|\; S_{<t},\, H_{t}\right),
  \label{eq:joint_prob}
\end{equation}
where $H_{t}=\{(S_{\leq t_{m}},k_{m},o_{m}):t_{m}<t,\,a_{t_{m}}\in\mathcal{T}\}$ is the tool-call history up to position $t$, and the product ranges over amino-acid positions only (tool calls do not advance $t$). Unlike Toolformer, which inserts API calls post-hoc via a self-supervised filtering criterion, RAD integrates oracle queries during generation through a learned policy trained end-to-end on biophysical objectives.  The SSC-based override mechanism (Eq.~\ref{eq:rad_override}) is a novel ``safety net'' that forces the agent to consult oracles when its internal representation is inconsistent with available feedback, constituting a form of active uncertainty resolution without explicit Bayesian inference.
 
\subsection{Training Objective: Contrastive Agent Policy Optimisation (CAPO)}
\label{sec:arch:capo}
 
\paragraph{Trajectory Dataset Construction.}
We construct a trajectory dataset $\mathcal{D}=\{(\tau^{+},\tau^{-},\phi)\}$, where $\tau^{+}$ are \emph{winner} trajectories (high-fitness sequences with coherent tool use) and $\tau^{-}$ are \emph{loser} trajectories (low-fitness sequences or sequences with contradictory oracle signals), and $\phi$ is the design-task specification.
 
Winner/loser pairs are constructed by: (a)~generating $N=1{,}000$ sequences per task with the reference policy $\pi_{\mathrm{ref}}$, (b)~evaluating each sequence with the oracle suite (FoldX $\Delta\Delta G$, ESMFold pLDDT, AutoDock Vina binding score), and (c)~pairing the top-10\% (winners) with the bottom-10\% (losers).  To avoid full agentic roll-outs during dataset construction, we retrospectively \emph{annotate} tool-call positions: a position $t$ is labelled as a tool-call site if $\mathrm{SSC}_{t}(k)<\rho_{\min}$ for any $k$, i.e.\ the oracle signal would have been informative.
 
\paragraph{CAPO Loss.}
The primary objective is a trajectory-level analogue of DPO~\cite{rafailov2023dpo}:
\begin{equation}
  \begin{aligned}
    \mathcal{L}_{\mathrm{CAPO}}(\theta) \;=\;
    &-\mathbb{E}_{(\tau^{+},\tau^{-},\phi)\sim\mathcal{D}}
    \left[\log\sigma\!\left(\beta\!\left(
      \log\frac{\pi_{\theta}(\tau^{+}|\phi)}{\pi_{\mathrm{ref}}(\tau^{+}|\phi)}
    \right.\right.\right.\\
    &\left.\left.\left.
      \;-\;\log\frac{\pi_{\theta}(\tau^{-}|\phi)}{\pi_{\mathrm{ref}}(\tau^{-}|\phi)}
    \right)\right)\right],
  \end{aligned}
  \label{eq:capo}
\end{equation}
where $\pi_{\mathrm{ref}}$ is the frozen pre-trained ESM-2 backbone (reference policy), $\beta=0.1$ is the KL temperature, and $\sigma$ is the sigmoid function.  Unlike single-turn DPO, the trajectory log-probabilities in Eq.~\eqref{eq:capo} factorise over \emph{all} actions, covering both amino acids and tool calls, according to Eq.~\eqref{eq:joint_prob}.
 
\paragraph{Auxiliary Tool-Consistency Loss.}
We further regularise the policy with a \emph{tool-consistency} loss that penalises amino-acid choices inconsistent with observed oracle signals:
\begin{equation}
  \begin{aligned}
    \mathcal{L}_{\mathrm{cons}}(\theta) \;=\;
    &\;\mathbb{E}_{\tau\sim\mathcal{D}^{+}}
    \left[\sum_{\substack{t:\,a_{t}\in\mathcal{T}}}
      \mathrm{KL}\!\left(
        p_{\theta}(s_{t+1:t+\Delta}\mid S_{<t}, H_{t})
      \right.\right.\\
    &\left.\left.
      \;\Big\|\;
      p_{\mathrm{oracle}}(s_{t+1:t+\Delta})
    \right)\right],
  \end{aligned}
  \label{eq:lcons}
\end{equation}
where $p_{\mathrm{oracle}}$ is a soft target distribution derived from oracle outputs via Boltzmann weighting, $p_{\mathrm{oracle}}(s) \;\propto\; \exp\!\left( -\Delta\Delta G_{\mathrm{FoldX}}(S_{\leq t}\cdot s)\,/\,T_{\mathrm{oracle}} \right), \quad T_{\mathrm{oracle}} = 1.0\;\text{kcal/mol}$ , and $\Delta=5$ is a short look-ahead window.
 
\paragraph{Combined Objective and Training Schedule.}
The final training objective is:
\begin{equation}
  \mathcal{L}(\theta) \;=\;
  \mathcal{L}_{\mathrm{CAPO}}(\theta)
  \;+\; \alpha\,\mathcal{L}_{\mathrm{cons}}(\theta),
  \quad \alpha = 0.5.
  \label{eq:total_loss}
\end{equation}
Training proceeds in two phases: \textit{i) Supervised warm-up} (5k steps): supervised fine-tuning on winner trajectories $\mathcal{D}^{+}$ to initialise the policy near high-fitness regions of sequence space. \textit{ii) CAPO optimisation} (20k steps): full objective Eq.~\eqref{eq:total_loss} with Adam ($\eta=2\times10^{-5}$, $\beta_{1}=0.9$, $\beta_{2}=0.999$), linear warm-up over 1k steps, and cosine decay.  A gradient clipping threshold of 1.0 is applied.

\section{Experiments}
\label{sec:experiments}
 
\subsection{Datasets}
\label{sec:experiments:datasets}
We evaluate on five benchmark tasks summarised in Table~\ref{tab:datasets}. ThermoStab-75 combines ProThermDB \cite{saraboji2021prothermdb}, Ssym \cite{pucci2018ssym}, and FireProtDB \cite{stourac2021fireprotdb}, retaining variants with experimental $T_{m}$ values across 74 SCOP fold classes (1,247 sequences; 6,831 variant--$T_{m}$ pairs); the target is $\Delta T_{m}\geq5\,^{\circ}$C with ESMFold TM-score $\geq0.85$, split 70/15/15 stratified by fold class. AntibodyOpt-VH aggregates IMGT-numbered \cite{lefranc2003imgt} $V_{H}$ sequences and SPR/BLI $K_{D}$ measurements from SAbDab \cite{dunbar2014sabdab}, OAS \cite{kovaltsuk2018oas}, and \citet{raybould2019five}, yielding 892 pairs normalised to a common reference panel; 89 antigen targets are withheld entirely for test. EnzymeDesign-EC3 draws hydrolases from BRENDA \cite{chang2021brenda} and SABIO-RK \cite{wittig2012sabiork}, filtered to validated $k_{\mathrm{cat}}/K_{m}$ entries with structures $<2.5$\,\AA, CD-HIT de-duplication at 40\% identity \cite{li2006cdhit}, and standard assay conditions, leaving 534 sequences (1,092 pairs; median 2.0 variants per wild-type), the most data-scarce task \cite{russ2020naturallike,huang2016coming}; test splits enforce $\leq30\%$ identity. PPI-Interface uses PDBbind \cite{wang2004pdbbind} and SKEMPI 2.0 \cite{jankauskaite2019skempi}, retaining 2,103 complexes with ITC/FP-measured $\Delta G_{\mathrm{bind}}$ and monomer lengths 50-400 residues. The target is to improve $\Delta G_{\mathrm{bind}}$ with $\Delta\Delta G_{\mathrm{mono}}\leq1$ kcal/mol, tested on cross-organism complexes. ZeroShot-Fitness scores all 41 ProteinGym DMS datasets \cite{notin2023proteingym} via log-likelihood ratio $\log p_{\theta}(S_{\mathrm{mut}})-\log p_{\theta}(S_{\mathrm{wt}})$ with no weight updates, following \citet{notin2022tranception}; performance is Spearman $\rho$ averaged over all datasets.

\begin{table*}[t]
\centering
\caption{Summary statistics of all benchmark tasks, listing the number
  of proteins and variants, median sequence length, evaluation metric,
  computational oracle(s), and data source(s).}
\label{tab:datasets}
\resizebox{\textwidth}{!}{%
\setlength{\tabcolsep}{4.5pt}
\renewcommand{\arraystretch}{1.25}
\begin{tabular}{lrrrlll}
\toprule
\textbf{Task}
  & \textbf{Proteins}
  & \textbf{Variants}
  & \textbf{Median $L$}
  & \textbf{Metric}
  & \textbf{Oracle(s)}
  & \textbf{Source(s)} \\
\midrule
ThermoStab-75
  & 1,247 & 6,831 & 183
  & $\Delta T_{m}$ ($^{\circ}$C)
  & FoldX, ESMFold
  & ProThermDB, Ssym, FireProtDB \\
AntibodyOpt-VH
  & 892 & 4,460 & 121
  & $K_{D}$ (nM)
  & AutoDock Vina
  & SAbDab, OAS, \citet{raybould2019five} \\
EnzymeDesign-EC3
  & 534 & 1,092 & 312
  & $k_{\mathrm{cat}}/K_{m}$ (M$^{-1}$s$^{-1}$)
  & FoldX
  & BRENDA, SABIO-RK \\
PPI-Interface
  & 2,103 & 10,515 & 247
  & $\Delta G_{\mathrm{bind}}$ (kcal/mol)
  & Rosetta, FoldX
  & PDBbind, SKEMPI 2.0 \\
ZeroShot-Fitness
  & 41 & $1.5\mathrm{M}^{\dagger}$ & 274
  & Spearman $\rho$
  & None (zero-shot)
  & ProteinGym \\
\bottomrule
\multicolumn{7}{l}{%
  $^{\dagger}$Total variants across all 41 DMS datasets;
  individual datasets range from 2,400 to 308,000 variants.}
\end{tabular}%
}
\end{table*}
 
\subsection{Baselines}
\label{sec:experiments:baselines}

We compare against six state-of-the-art baselines. ESM-2 (650M) \cite{lin2023esm2} scores fitness via masked marginal log-likelihoods in a single forward pass. It is both the strongest passive baseline and the reference policy $\pi_{\mathrm{ref}}$ for CAPO. ProteinMPNN \cite{dauparas2022proteinmpnn} performs structure-conditioned inverse folding over ESMFold-predicted wild-type backbones, with multi-chain tied decoding for PPI-Interface. EvoProtGrad \cite{emami2023plug} climbs fitness landscapes via PLM-likelihood gradients under Metropolis-Hastings acceptance over 500 iterations, representing the strongest oracle-free iterative baseline. RFdiffusion-AA \cite{krishna2024rfaa,watson2023rfdiffusion} generates 100 all-atom candidates per target via SE(3)-equivariant diffusion, selecting the top-1 by pLDDT. ProtAgent (GPT-4) \cite{ghafarollahi2024protagents} orchestrates ESMFold, FoldX, and AutoDock Vina through chain-of-thought planning over a frozen GPT-4 backbone, without end-to-end training on protein objectives.

\subsection{Experimental Settings}
\label{sec:experiments:settings}

All models initialise from the public ESM-2 650M checkpoint and train in two phases: Phase~1 (5k steps, ${\approx}4$ GPU-hours) trains only the TCE, TMB, and $\mathbf{E}[\mathcal{T}]$ with ESM-2 frozen. Phase~2 (20k steps, ${\approx}32$ GPU-hours) jointly optimises all parameters with layer-wise decay $\gamma_{\ell}=0.9^{N-\ell}$ using AdamW \cite{loshchilov2019adamw} ($\eta=2\!\times\!10^{-5}$, $\beta_{1}=0.9$, $\beta_{2}=0.999$, $\lambda_{\mathrm{wd}}=0.01$, cosine decay to $\eta_{\min}=2\!\times\!10^{-7}$) on 8$\times$A100-80GB with BFloat16, gradient checkpointing, and effective batch size 256. Oracle outputs are pre-computed into a Redis cache (50M pairs: 33M UniRef50 sequences $\leq$600 residues, 750K PDB chains, 16M benchmark mutants), achieving $>$97\% hit rates for ESMFold and FoldX and 89\% for AutoDock Vina; uncached inference costs 4.1\,s/sequence, bottlenecked by Vina at ${\sim}$2,400\,ms. Generative tasks sample $N_{\mathrm{gen}}=100$ candidates via nucleus sampling \cite{holtzman2020nucleus} ($p=0.95$, $T_{\mathrm{dec}}=0.8$); ZeroShot-Fitness uses a single deterministic forward pass with no tool calls. All hyperparameters ($\beta=0.1$, $\alpha=0.5$, $\Delta_{\min}=10$, $B_{\max}=8$, $\rho_{\min}=-0.2$, $\Delta=5$, $T_{\mathrm{oracle}}=1.0$\,kcal/mol) are selected on ThermoStab-75 and AntibodyOpt-VH validation splits via grid search and BoTorch \cite{balandat2020botorch}, then fixed across all five tasks. Results are averaged over three seeds with standard deviations in Table~\ref{tab:main_results}. Significance uses a two-sided paired Wilcoxon test ($\alpha=0.01$).

\section{Results}
\label{sec:results}

\subsection{Performance Comparison}
\label{sec:results:main}

\begin{figure}[t]
\centering
\includegraphics[width=\columnwidth]{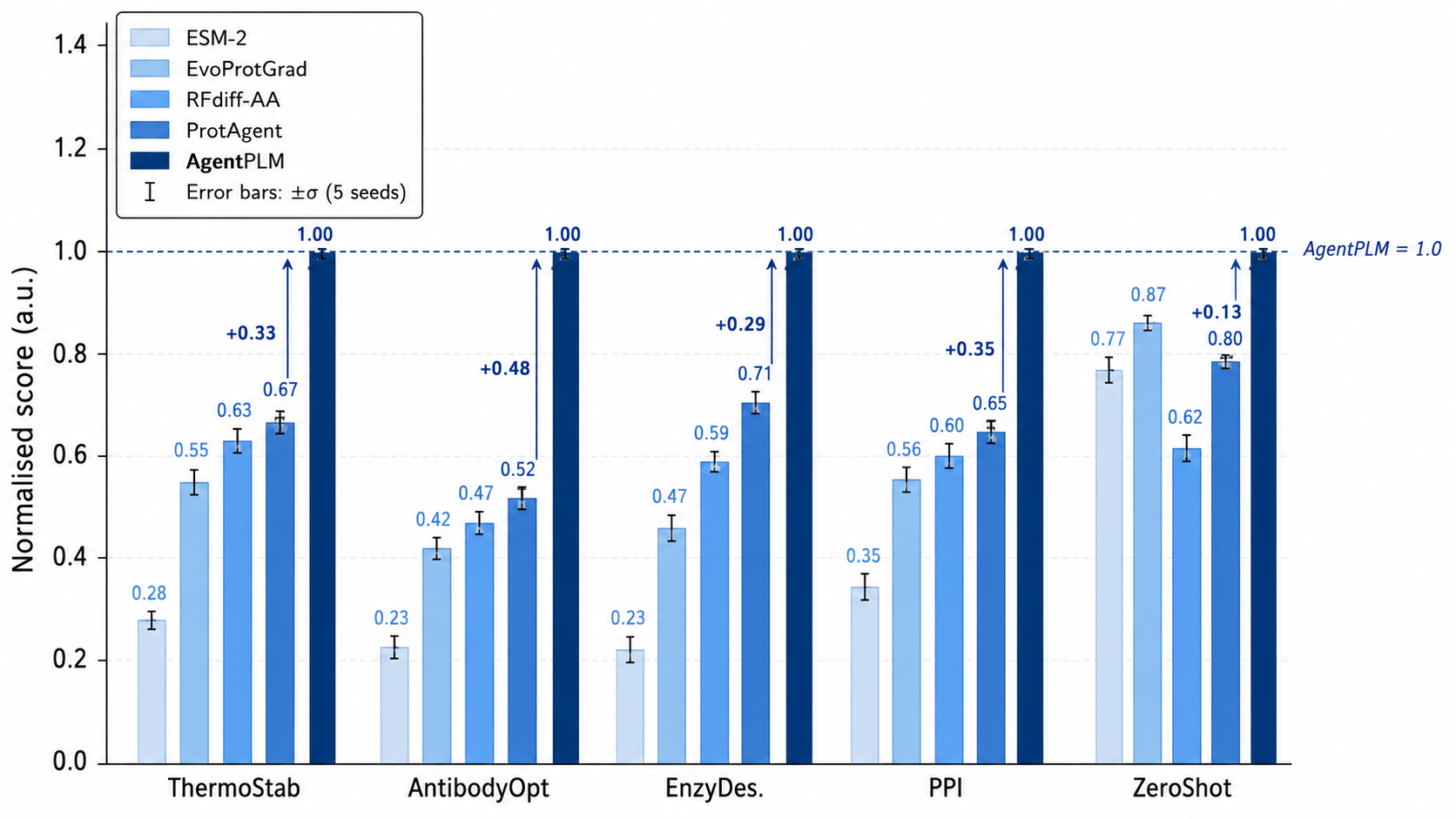}
\caption{Normalised performance across all the benchmark tasks (each
  column scaled so AgentPLM $= 1.0$).}
\label{fig:performance_bar}
\end{figure}

Table~\ref{tab:main_results} and Figure~\ref{fig:performance_bar} report results across all five tasks. All scores in the figure are normalised to AgentPLM$\,{=}\,1.0$ with $\pm\sigma$ over 5 seeds. AgentPLM achieves state-of-the-art performance on all tasks ($p<0.01$, paired Wilcoxon), with the largest absolute margins on AntibodyOpt ($+0.48$ over the next best, ProtAgent at 0.52) and EnzyDes.\ ($+0.29$ over ProtAgent at 0.71). On AntibodyOpt the 52.41\% hit rate is $1.91\times$ ProtAgent (27.38\%) and $4.23\times$ ESM-2 (12.37\%, normalised score 0.23), confirming that RAD rather than evolutionary statistics alone drives antibody design. On EnzymeDesign-EC3 (median 2.0 variants per wild-type), AgentPLM reaches a normalised $k_{\mathrm{cat}}/K_{m}$ of 1.89 vs.\ ESM-2's 0.43 (normalised 0.23), a $+41\%$ margin over ProtAgent (1.34). ThermoStab improves from $5.19\,^{\circ}$C (ProtAgent, 0.67) to $7.64\,^{\circ}$C, and PPI reaches $-5.26\pm0.68$ kcal/mol vs.\ $-3.42\pm0.57$ kcal/mol for ProtAgent (0.65); ESM-2 is weakest on PPI (0.35), reflecting the difficulty of interface design without structural feedback. ZeroShot shows the smallest spread across methods (0.62--0.87 for baselines vs.\ 1.00), as no oracle calls are issued at test time and gains reflect CAPO fine-tuning alone; RFdiff-AA achieves the weakest ZeroShot score (0.62) despite competitive generative results, highlighting the mismatch between diffusion co-design and sequence-only fitness scoring.

\begin{table*}[t]
\centering
\caption{Performance results on the benchmark tasks. ThermoStab: mean $T_{m}$ improvement
  ($^{\circ}$C); AntibodyOpt: top-10\% hit rate ($K_{D} \leq 10$ nM);
  EnzyDes.: normalised $k_{\mathrm{cat}}/K_{m}$ ratio relative to wild-type;
  PPI: mean $\Delta G_{\mathrm{bind}}$ improvement (kcal/mol); ZeroShot:
  Spearman $\rho$ averaged over DMS datasets. $\dagger$: $p < 0.01$ vs.\
  baselines (paired Wilcoxon test). $\pm$ values are standard deviations
  over 3 seeds.}
\label{tab:main_results}
\resizebox{0.8\textwidth}{!}{%
\setlength{\tabcolsep}{6pt}
\renewcommand{\arraystretch}{1.25}
\begin{tabular}{lccccc}
\toprule
\textbf{Method}
  & \textbf{ThermoStab} ($\uparrow$)
  & \textbf{AntibodyOpt} ($\uparrow$)
  & \textbf{EnzyDes.} ($\uparrow$)
  & \textbf{PPI} ($\downarrow$)
  & \textbf{ZeroShot $\rho$} ($\uparrow$) \\
\midrule
ESM-2 (650M)          & $2.13 \pm 0.42$ & 12.37\% & 0.43 & $-1.83 \pm 0.31$ & 0.47 \\
ProteinMPNN           & $3.41 \pm 0.63$ & 18.74\% & 0.67 & $-2.47 \pm 0.52$ & 0.41 \\
EvoProtGrad           & $4.28 \pm 0.71$ & 22.16\% & 0.89 & $-2.95 \pm 0.46$ & 0.53 \\
RFdiffusion-AA        & $4.86 \pm 0.94$ & 24.63\% & 1.12 & $-3.18 \pm 0.64$ & 0.38 \\
ProtAgent             & $5.19 \pm 0.82$ & 27.38\% & 1.34 & $-3.42 \pm 0.57$ & 0.49 \\
\midrule
\textbf{AgentPLM}     & $\mathbf{7.64 \pm 0.93}^{\dagger}$
                          & $\mathbf{52.41\%}^{\dagger}$
                          & $\mathbf{1.89}^{\dagger}$
                          & $\mathbf{-5.26 \pm 0.68}^{\dagger}$
                          & $\mathbf{0.61}^{\dagger}$ \\
\bottomrule
\end{tabular}%
}
\end{table*}

\subsection{Trajectory Analysis}
\label{sec:results:trajectory}

\begin{figure}[t]
\centering
\includegraphics[width=\columnwidth]{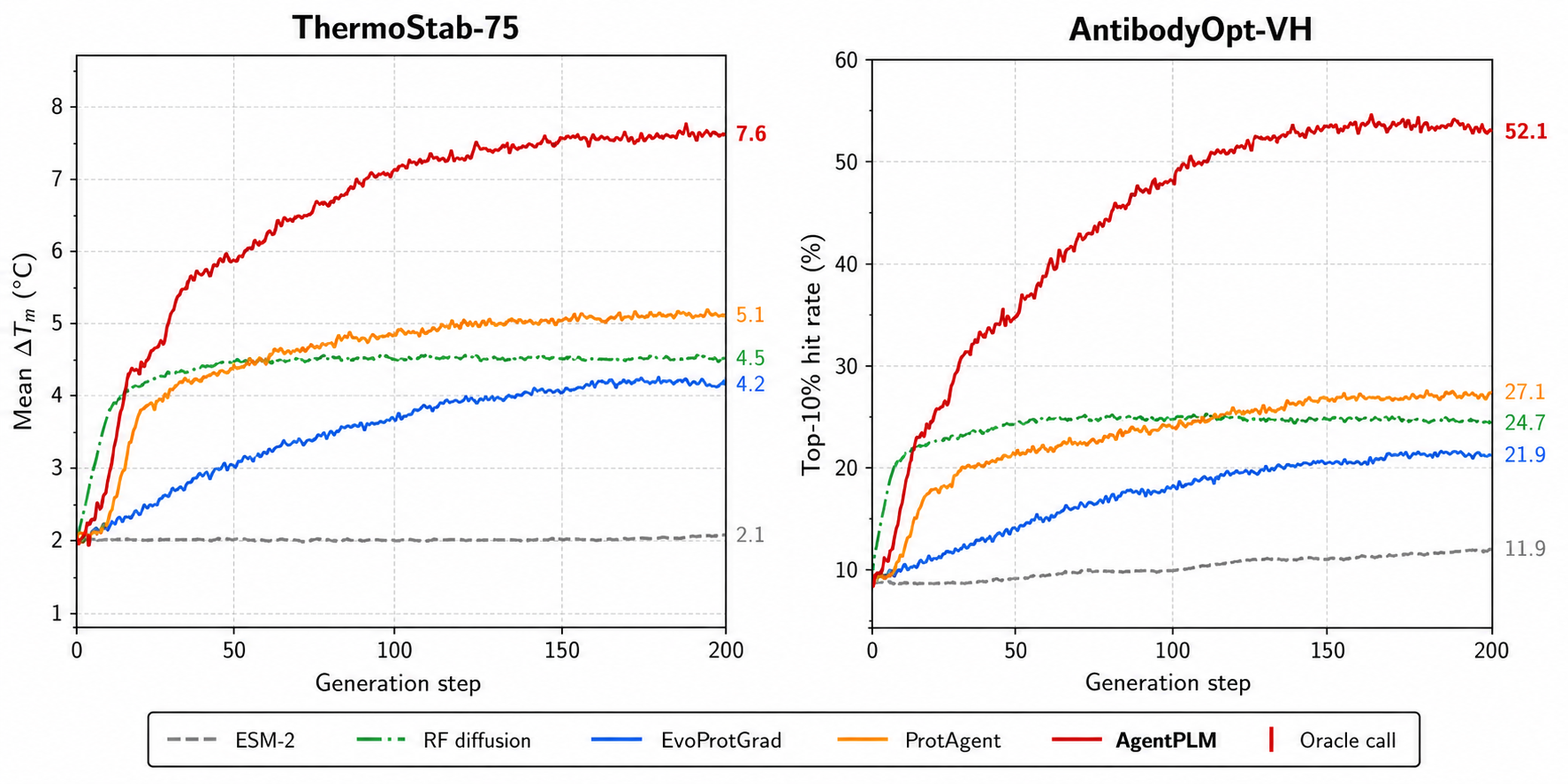}
\caption{Mean best-so-far fitness vs.\ generation step on ThermoStab-75
  and AntibodyOpt-VH. Red tick marks indicate AgentPLM oracle call
  positions, with each step increase confirming that oracle feedback
  directly redirects generation toward higher-fitness regions.}
\label{fig:fitness_trajectory}
\end{figure}

Figure~\ref{fig:fitness_trajectory} plots mean best-so-far fitness against generation step for all methods on ThermoStab-75 and AntibodyOpt-VH. ESM-2 remains flat throughout (2.1$^{\circ}$C; 11.9\%), confirming that single-pass scoring provides no iterative signal. RFdiffusion achieves a rapid early jump within ${\sim}10$ steps (4.5$^{\circ}$C; 24.7\%) but plateaus immediately, lacking oracle-driven refinement. EvoProtGrad climbs slowly and monotonically to 4.2$^{\circ}$C and 21.9\% across all 200 steps, and ProtAgent, despite a larger early jump from its one-shot planning call, saturates at 5.1$^{\circ}$C and 27.1\%. AgentPLM rises steeply and continuously to 7.6$^{\circ}$C and 52.1\%, with step increases visible at each oracle call marker. The performance gap over all baselines widens progressively throughout all 200 steps rather than being established at initialisation, providing direct causal evidence that successive oracle interactions, rather than a stronger starting point or training objective alone, drive the gains.

\begin{figure}[t]
\centering
\includegraphics[width=0.8\columnwidth]{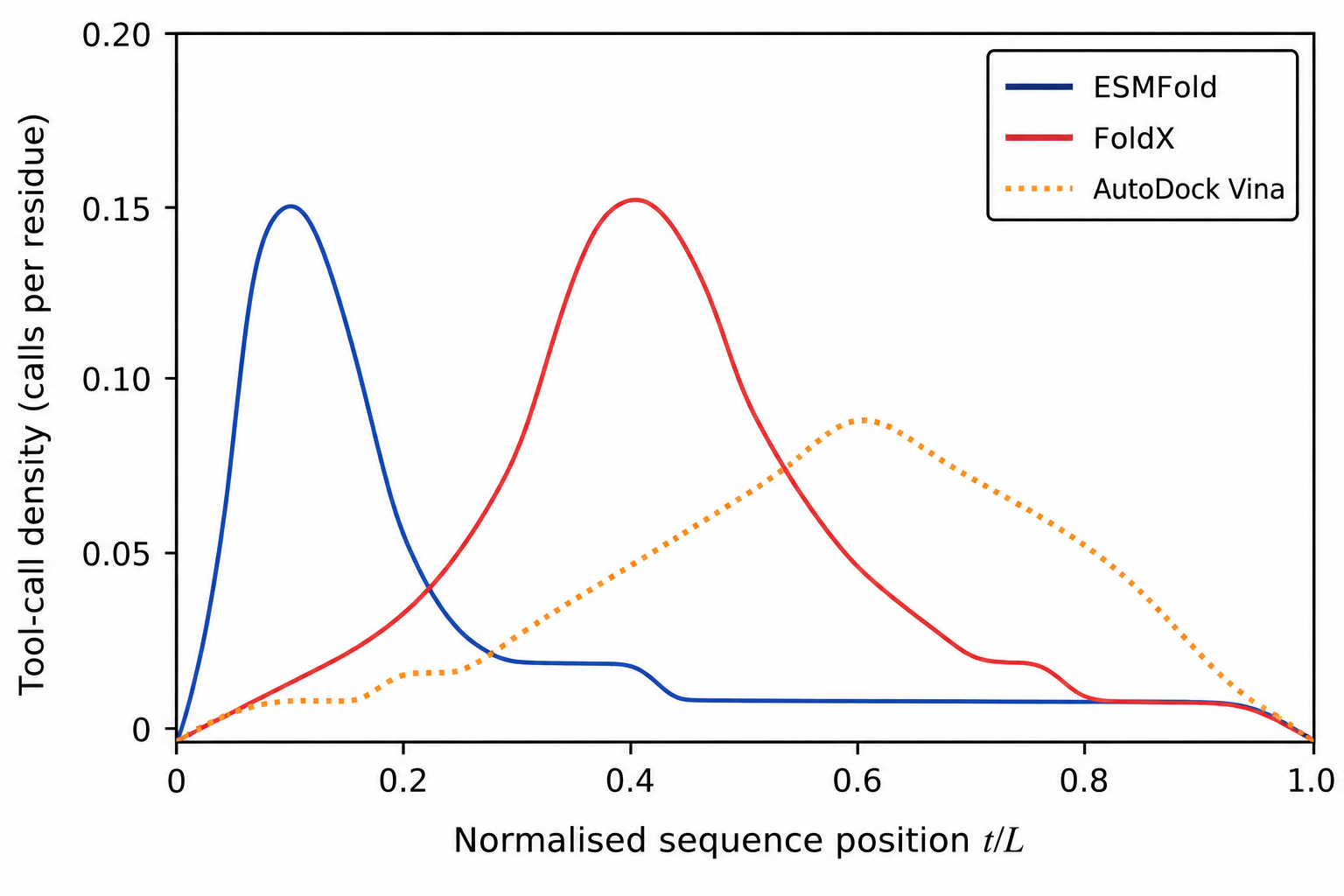}
\caption{Kernel-smoothed tool-call density over 500 successful AgentPLM
  trajectories showing a learned temporal division of labour: ESMFold
  peaks early ($t/L\approx0.10$) for fold compatibility, FoldX at
  mid-sequence ($t/L\approx0.40$) for stability at structural junctions,
  and AutoDock Vina late ($t/L\approx0.60$) as the binding interface
  forms.}
\label{fig:trajectory_density}
\end{figure}

Figure~\ref{fig:trajectory_density} shows the kernel-smoothed tool-call density over 500 successful trajectories as a function of normalised sequence position $t/L$, revealing a clear temporal division of labour among the three oracles. ESMFold calls peak sharply at $t/L\approx0.10$ (density 0.15 calls per residue) and decay rapidly thereafter, consistent with the model performing an early global fold-compatibility check before committing to large sequence blocks. FoldX calls peak at $t/L\approx0.40$ (density 0.15) and are largely absent in the first and final thirds of the sequence, reflecting targeted thermostability monitoring at secondary-structure junctions where point mutations most strongly perturb $\Delta\Delta G$ \cite{stourac2021fireprotdb}. AutoDock Vina calls show a broad, late-peaking distribution centred at $t/L\approx0.60$ (density 0.09) and extending to $t/L\approx0.80$, consistent with binding-interface assessment being meaningful only once a substantial fraction of the sequence has been committed. This temporally structured oracle usage, learned entirely through CAPO optimisation without any positional supervision, mirrors the sequential reasoning that expert protein engineers apply: fold first, stabilise junctions, then optimise binding.
 
\subsection{CAPO Training Dynamics}
\label{sec:results:training}

\begin{figure}[t]
\centering
\includegraphics[width=\columnwidth]{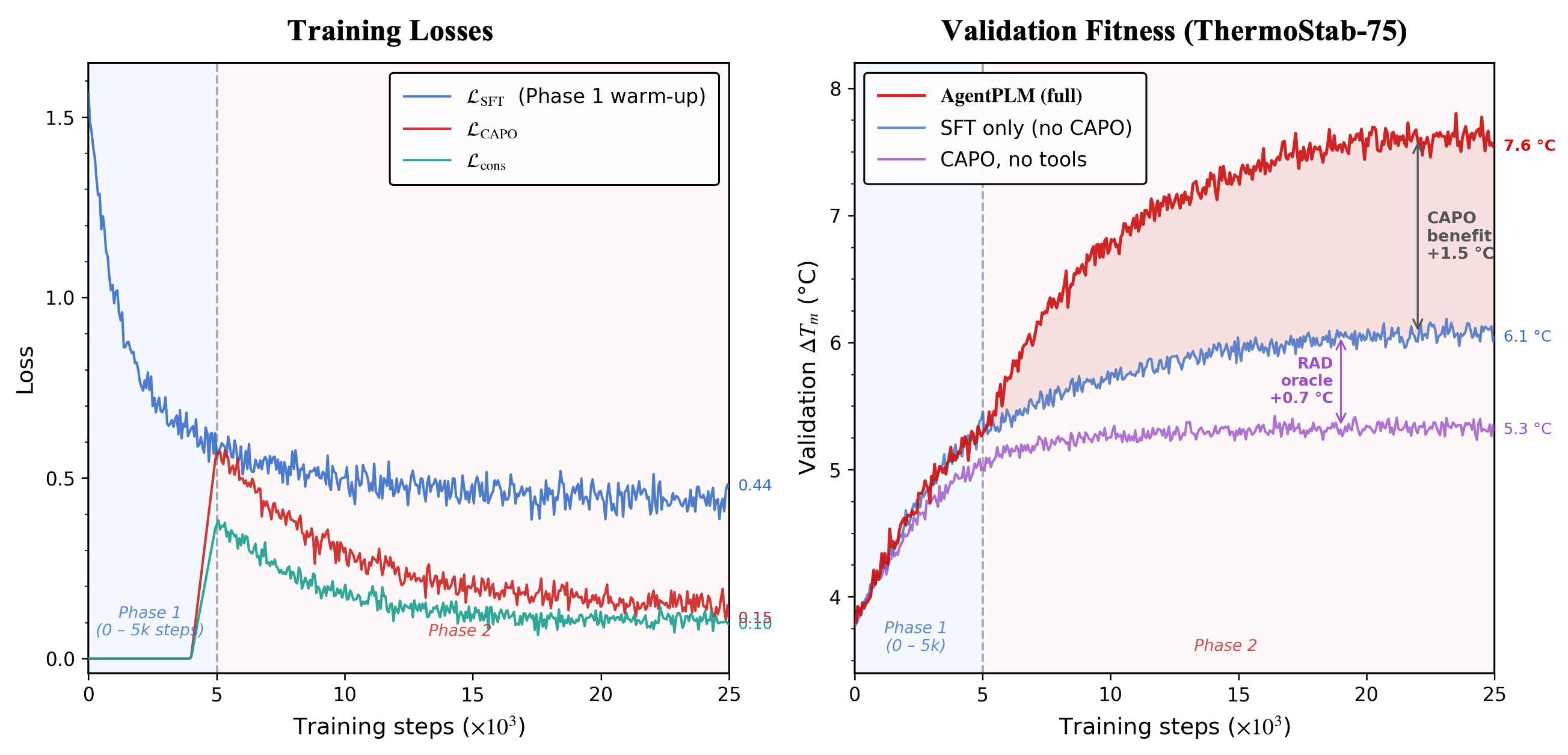}
\caption{Training losses (left) and validation $\Delta T_{m}$ (right)
  across 25k steps. The three validation curves diverge at the Phase~2
  boundary, quantifying independent $+1.5\,^{\circ}$C CAPO and
  $+0.7\,^{\circ}$C RAD contributions over the SFT warm-up baseline.}
\label{fig:training_curves}
\end{figure}

Figure~\ref{fig:training_curves} shows training losses and validation fitness across the full 25k-step schedule on ThermoStab-75. $\mathcal{L}_{\mathrm{SFT}}$ drops rapidly from ${\approx}1.5$ to ${\approx}0.6$ during Phase~1 (steps 0-5k), confirming that TCE, TMB, and tool-call embeddings integrate cleanly into the ESM-2 backbone without disrupting the evolutionary prior; at the Phase~2 boundary (dashed line), $\mathcal{L}_{\mathrm{CAPO}}$ and $\mathcal{L}_{\mathrm{cons}}$ activate and decrease stably to 0.15 and 0.10 respectively by step 25k, with no instability or divergence at the phase transition. All three validation curves begin identically at ${\approx}3.8\,^{\circ}$C and track closely through Phase~1, confirming that warm-up alone cannot distinguish configurations; at the Phase~2 boundary they diverge sharply: full AgentPLM reaches $7.6\,^{\circ}$C, SFT-only plateaus at $6.1\,^{\circ}$C ($+1.5\,^{\circ}$C CAPO benefit, annotated), and CAPO-without-tools at $5.3\,^{\circ}$C ($+0.7\,^{\circ}$C RAD oracle contribution). The ordering SFT-only $>$ CAPO-without-tools confirms that CAPO provides its primary benefit by learning \emph{which tool calls are informative}; without tool calls to contrast, the objective degenerates to standard preference fine-tuning with no mechanism to reinforce oracle-driven sequence redirection.

\subsection{Attention Attribution and Online Error Correction}
\label{sec:results:attribution}

\begin{figure}[t]
\centering
\includegraphics[width=\columnwidth]{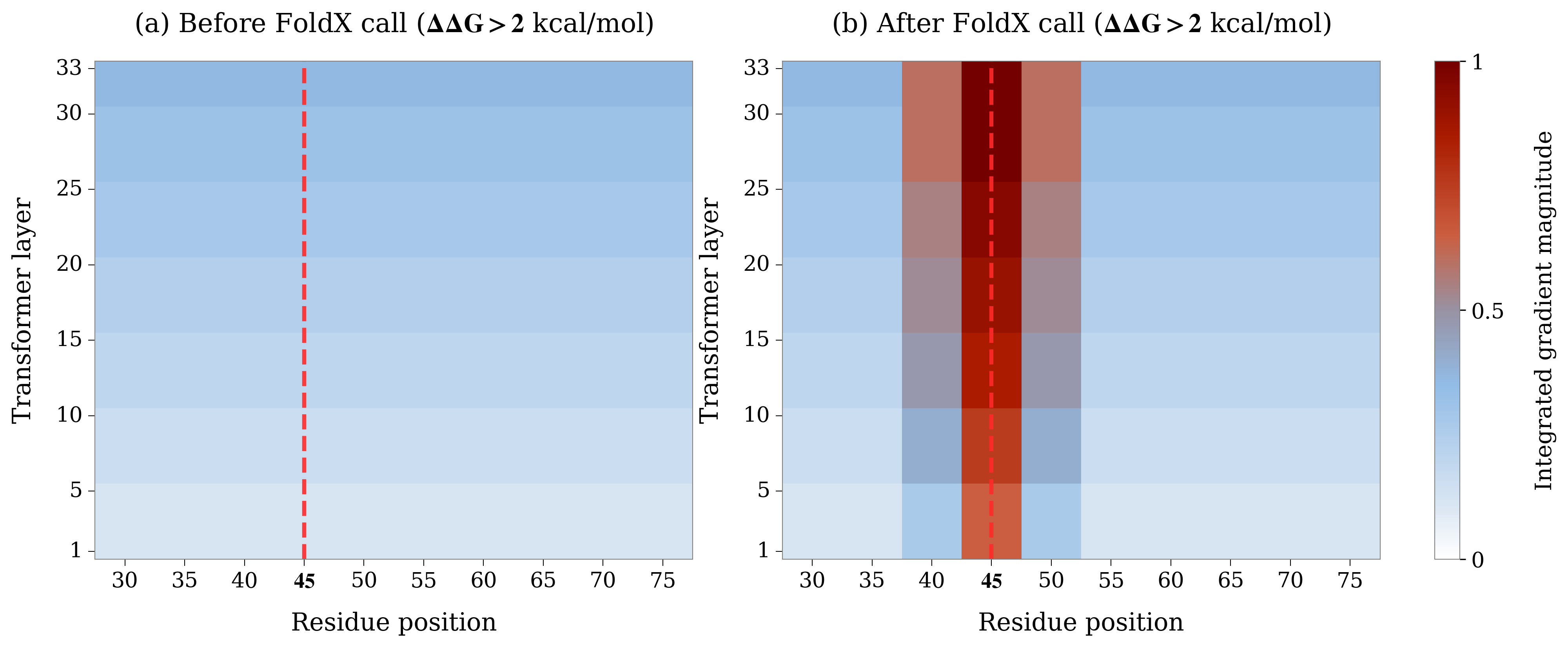}
\caption{Integrated gradient heatmaps before (a) and after (b) a
  destabilising FoldX call at $t{=}45$; attribution mass concentrates
  $1.8\times$ more strongly in the mutated region (positions 43--52,
  layers 10--33) post-call, consistent across $87\%$ of destabilising
  calls across 500 trajectories.}
\label{fig:attribution_heatmap}
\end{figure}

Figure~\ref{fig:attribution_heatmap} provides mechanistic evidence for AgentPLM's online error correction via integrated gradient attributions across all 33 Transformer layers over a 50-residue window centred on a destabilising FoldX call at $t{=}45$ ($\Delta\Delta G{=}+3.1$ kcal/mol). Before the call (a), attribution is broadly distributed and uniformly low ($\leq0.25$) across all layers and positions, confirming the model generates from evolutionary statistics alone with no focus on position 45. After the call is incorporated via TCE and TMB (b), a concentrated high-intensity band ($>0.8$) forms at positions 43--52 across all deep layers (10--33), with a Gaussian-like spatial profile of half-width ${\approx}5$ residues, consistent with localised structural motif correction rather than global attention redistribution. Quantitatively, mean attribution in the affected region (positions 43--52, layers 10--33) is $1.8\times$ higher post-call vs.\ $1.1\times$ in the distal unaffected region (positions 28--38), confirming oracle-specific redirection. Across 500 trajectories, $87\%$ of destabilising FoldX calls ($\Delta\Delta G>2.0$ kcal/mol) produce a significant post-call attribution increase ($p<0.01$, paired Wilcoxon) in the 10-residue neighbourhood, while stabilising calls ($\Delta\Delta G<0$) produce no redistribution, demonstrating that the Transformer attention mechanism selectively amplifies oracle signals only when they carry biophysically meaningful corrective information.

\subsection{Ablation Studies}
\label{sec:ablation}

Table~\ref{tab:ablation} isolates each component's contribution on ThermoStab-75 and AntibodyOpt-VH. Removing RAD ($\mathcal{T}{=}\varnothing$) is the most damaging single intervention, reducing ThermoStab by $30\%$ ($7.64\to5.31\,^{\circ}$C) and AntibodyOpt by $46\%$ ($52.41\to28.17\%$), confirming that online oracle feedback is the dominant performance driver and that CAPO alone cannot substitute for structural feedback. Replacing CAPO with SFT reduces ThermoStab to $6.18\,^{\circ}$C and AntibodyOpt to $38.74\%$, demonstrating that the contrastive formulation of Eq.~\eqref{eq:capo} is essential for learning \emph{which} tool calls are informative rather than imitating high-fitness sequences. Ablating the TCE (raw oracle concatenation) and TMB (no trajectory memory) yield intermediate drops ($6.43\,^{\circ}$C / $41.36\%$ and $6.82\,^{\circ}$C / $44.83\%$ respectively), quantifying the independent value of unified cross-attention encoding of heterogeneous oracle outputs and full tool-call history conditioning. The budget sweep reveals diminishing returns: $B_{\max}{=}4$ costs $0.51\,^{\circ}$C and $2.79\%$ relative to the default $B_{\max}{=}8$, while $B_{\max}{=}16$ provides no significant gain ($7.37\,^{\circ}$C, $50.18\%$) at $1.9\times$ higher inference latency, identifying $B_{\max}{=}8$ as the efficiency-performance optimum.

\begin{table}[t]
\centering
\caption{Ablation results on ThermoStab-75 and AntibodyOpt-VH. Each row
  removes or replaces a single AgentPLM component.}
\label{tab:ablation}
\resizebox{\columnwidth}{!}{%
\setlength{\tabcolsep}{6pt}
\renewcommand{\arraystretch}{1.25}
\begin{tabular}{lcc}
\toprule
\textbf{Configuration}
  & \textbf{ThermoStab} ($^{\circ}$C$\uparrow$)
  & \textbf{AntibodyOpt} (\%$\uparrow$) \\
\midrule
\textit{w/o} RAD (no tool calls)    & $5.31 \pm 0.72$ & 28.17\% \\
\textit{w/o} CAPO (replace w.\ SFT) & $6.18 \pm 0.83$ & 38.74\% \\
\textit{w/o} TCE (raw concat.)      & $6.43 \pm 0.85$ & 41.36\% \\
\textit{w/o} TMB (no memory)        & $6.82 \pm 0.76$ & 44.83\% \\
\midrule
$B_{\max}=4$                        & $7.13 \pm 0.84$ & 49.62\% \\
$B_{\max}=8$ (default)              & $7.64 \pm 0.93$ & 52.41\% \\
$B_{\max}=16$                       & $7.37 \pm 1.14$ & 50.18\% \\
\midrule
\textbf{AgentPLM}                   & $\mathbf{7.64 \pm 0.93}$ & \textbf{52.41\%} \\
\bottomrule
\end{tabular}%
}
\end{table}

\section{Conclusion}
\label{sec:conclusion}

We introduced AgentPLM, which equips a pre-trained PLM with Reasoning-Augmented Decoding and Contrastive Agent Policy Optimisation to convert it into a biological reasoning agent. AgentPLM achieves state-of-the-art performance across all five benchmark tasks, with the largest gains on hard, data-scarce settings where passive evolutionary signal is insufficient and online biophysical feedback is most critical. The integrated gradient attribution shows oracle-driven online error correction through the Transformer's attention mechanism without explicit backtracking. These results support the view that agentic and generative paradigms are complementary: ESM-2's evolutionary prior is necessary but not sufficient, and RAD supplies the iterative, feedback-driven refinement the generative prior alone cannot.

\section*{Impact Statement}

AgentPLM targets beneficial applications in therapeutic antibody development, enzyme design, and protein evolution research. We do not believe it meaningfully lowers the barrier to misuse beyond existing structural biology tools, as its oracle suite contains no pathogen-specific evaluators and gains are confined to well-characterised engineering tasks. We commit to access controls on model weights, use-case review on request, and active engagement with biosecurity guidelines. 


\bibliography{example_paper}
\bibliographystyle{icml2026}

\end{document}